\documentclass{article}

\usepackage{microtype}
\usepackage{graphicx}
\usepackage{subfigure}
\usepackage{booktabs} 
\usepackage{amsfonts}
\usepackage{amsmath}
\usepackage{amssymb}

\graphicspath{ {./images/} }
\usepackage{hyperref}

\DeclareMathOperator{\sgn}{sgn}
\DeclareMathOperator{\diag}{diag}
\DeclareMathOperator{\quantile}{quantile}
\DeclareMathOperator{\Uniform}{Uniform}

\usepackage[accepted]{icml2020}

\icmltitlerunning{ZeroGrad : Mitigating and Explaining Catastrophic Overfitting in FGSM Adversarial Training}

\begin{document}

\twocolumn[
\icmltitle{ZeroGrad : Mitigating and Explaining Catastrophic Overfitting in FGSM Adversarial Training}



\icmlsetsymbol{equal}{*}

\begin{icmlauthorlist}
\icmlauthor{Zeinab Golgooni}{equal,ce}
\icmlauthor{Mehrdad Saberi}{equal,ce}
\icmlauthor{Masih Eskandar}{equal,ce}
\icmlauthor{Mohammad Hossein Rohban}{ce}

\end{icmlauthorlist}
\icmlaffiliation{ce}{Department of Computer Engineering, Sharif University of Technology, Tehran, Iran}


\icmlcorrespondingauthor{Mohammad Hossein Rohban}{rohban@sharif.edu}

\icmlkeywords{Machine Learning, ICML}

\vskip 0.3in
]



\printAffiliationsAndNotice{\icmlEqualContribution} 

\begin{abstract}
Making deep neural networks robust to small adversarial noises has recently been sought in many applications. Adversarial training through iterative projected gradient descent (PGD) has been established as one of the mainstream ideas to achieve this goal. However, PGD is computationally demanding and often prohibitive in case of large datasets and models. For this reason, single-step PGD, also known as FGSM, has recently gained interest in the field. Unfortunately, FGSM-training leads to a phenomenon called ``catastrophic overfitting," which is a sudden drop in the adversarial accuracy under the PGD attack. In this paper, we support the idea that small input gradients play a key role in this phenomenon, and hence propose to zero the input gradient elements that are small for crafting FGSM attacks. Our proposed idea, while being simple and efficient, achieves competitive adversarial accuracy on various datasets. \footnote{Code is available at \url{https://github.com/rohban-lab/catastrophic_overfitting}}
\end{abstract}

\section{Introduction}
\label{intro}
Despite deep neural networks  impressive success in many real-world problems, they suffer from instability under test-time adversarial noise. Training the network based on the adversarial samples in each mini-batch, which is known as ``Adversarial training" (AT) \cite{madry2018}, has been empirically established as a general and effective approach to remedy this issue. Multi-step gradient-based maximization of the loss function with respect to the norm-bounded perturbations is often used to craft the adversarial samples in AT. This method is known as Projected Gradient Descent (PGD) attack and results in reasonably well generalization of the adversarially trained networks under many strong attacks \cite{carlini2017towards}, under the same/similar threat models. The iterative process, however, makes this method very slow and sometimes infeasible in case of large datasets and models. Fast Gradient Sign Method (FGSM) \cite{Goodfellow2015ExplainingAH}, which is the single-step PGD, is much faster but results in poor generalization under the stronger multi-step PGD attack. 

Recently, few attempts have been made to address the generalization issues of FGSM \cite{wong2020fast}. ``Fast," introduced FGSM-RS, an extended version of FGSM by adding a random uniform noise prior to applying FGSM. Although this method can noticeably improve vanilla FGSM-trained networks, it suffers from an unknown phenomenon called “catastrophic overfitting” \cite{wong2020fast}. This refers to a sudden and drastic drop of the adversarial test accuracy, under a stronger PGD attack, at a single epoch, although the training robust accuracy under FGSM continues to go up. That is, a large gap between FGSM and PGD losses can be seen after this epoch. At a first look, this event can be seen as an overfitting to the FGSM attack due to the weakness of FGSM attack \cite{wong2020fast}, but the suddenness of this failure warrants the need for deeper explanations. Following this track, understanding catastrophic overfitting, and improving the FGSM training has recently gained attention \cite{li2020towards, kim2020understanding, vivek2020single, vivek2020regularizers, andriushchenko2020understanding}. 

\cite{li2020towards} noticed that by inspecting the test robust accuracy at the mini-batch resolution, catastrophic overfitting happens even in successful runs of FGSM training. They hypothesized that the initial random noise in FGSM-RS serves as a way to recover from such drops in the model robustness, but because of the noise randomness, this recovery may fail with a non-zero probability. According to this hypothesis, they proposed to use PGD training as a better recovery mechanism whenever overfitting happens. They were able to mitigate the issue with only a 2-3 times training slowdown compared to training with FGSM, which is indeed fascinating. However, this method does not take a step toward explaining why the overfitting happens. Another downside is that the attacks like PGD are multi-step and cannot be parallelized to run as fast as single-step attacks on current computational units.

As another solution, \cite{vivek2020regularizers} proposed to regularize the difference in network logits for the FGSM and iterative-FGSM perturbed inputs to be close to zero in a few mini-batches. They have empirically shown that such a regularization could prevent gradient masking. On the other hand, \cite{kim2020understanding} takes another perspective and points out that due to FGSM perturbations always lying on the boundaries of the $\ell_\infty$ ball, the network loss faces the so-called ``boundary distortion." This distortion makes the loss surface highly curved, and consequently results in the model becoming non-robust against smaller perturbations. Therefore, they proposed to evaluate the loss at various multiples of the FGSM direction, where multiples are between 0 and 1 and use the smallest multiple that results in misclassification. 

Among the most insightful attempts, GradAlign \cite{andriushchenko2020understanding} hypothesized that large variations of the input gradient around a sample cause the FGSM direction to be noisy and irrelevant, and consequently result in poor attack quality and overfitting. They backed up their hypothesis by theoretically showing that the angle between local input gradients is upper-bounded prior to the training for an appropriate Gaussian weight initialization. This partly explains why it takes a few epochs before catastrophic overfitting occurs. Based on this explanation, they proposed to regularize the optimization by constraining such variations to be small. This proved to be effective in a lot of cases where FGSM fails. However, the mentioned regularization requires a sequential procedure called ``double-backpropagation," and increases the training time by a factor of 2-3 compared to the vanilla FGSM-training. Furthermore, neither of the explanations in earlier work fully characterize such an event, e.g. why such drops in the robust generalization occur so quickly. 

In this paper, we hypothesize and support that tiny input gradients play a key role in catastrophic overfitting. Our intuition is that due to the incontinuity of the ``sign" function in FGSM, elements of the input gradient that are close to zero could cause drastic change in the attack and large weight updates in two consecutive epochs. This tends to take place later in the training, as it has been empirically observed that adversarially robust models yield sparser input gradients, resulting in many close to zero gradient elements later in training. Based on this explanation, in crafting the FGSM attacks, we propose to zero out the input gradient elements whose absolute values are below a certain threshold. As an alternative solution, we propose to zero out gradient elements that have inconsistent signs across multiple random starts. We observe that such simple remedies prevent catastrophic overfitting and result in competitive robust accuracies. Our proposed scheme, as opposed to the prior work, does not have any significant train-time overhead, or could else be parallelized. In addition, we notice that based on our theoretical insight about large sudden weight updates, regularizing second derivative of the loss, which is implicitly done in GradAlign, could help to avoid overfitting. This also highlights the generality of our insights.

Our contributions are summarized below:
\begin{itemize}
    \item a more comprehensive explanation of catastrophic overfitting;
    \item two simple yet effective remedies based on our explanation to avoid this issue;
    \item competitive adversarial accuracy on various datasets under no significant train-time overhead, considering the possibility of parallelization.
\end{itemize}

\section{Causes of Catastrophic Overfitting}
\label{section-2}

In this section, we provide some theoretical insights into why catastrophic overfitting happens, and why it is sudden and is often observed in later epochs of training. We then back up our hypothesis with some numerical results. 

\subsection{Theoretical Insights}

Let $f(x, W)$ represent a classification hypothesis (e.g. a deep neural network) with adjustable parameters $W$ and input $x$, and $g(x, W) := \ell(f(x, W), y)$ be the loss function, with $y$ as the ground-truth label that is used in training (e.g. cross-entropy loss). Further let $\eta_i = \epsilon \sgn \nabla_x g(x_i, W)$ be the FGSM attack on the data point $x_i$. So $x^\prime_i = x_i + \alpha \eta_i$ represents the adversarial example by FGSM, where $\alpha$ is the FGSM step size.

We are implicitly solving the following optimizing during the FGSM training:
\begin{equation}
    \min_W \sum_{i} g(x^\prime_i, W)
\end{equation}
A necessary condition for the optimizer of this loss in the local minimum is the following:
\begin{equation} \label{Suff_cond}
\begin{split}
    \nabla_W & \sum_i g(x^\prime_i, W) \\
    & = \sum_{i} \alpha \left\{\nabla_W \eta_i\right\} \bigg\rvert_{(x_i, W)} .\left\{\nabla_x g \right\} \bigg\rvert_{(x^\prime_i, W)} \\ & ~~~~~~~~~~~~~ + \nabla_W g \bigg\rvert_{(x^\prime_i, W)} \\ & = {\bf 0}
\end{split}    
\end{equation}
But note that,
\begin{equation} \label{grad_eta}
    \nabla_W \eta_i = \epsilon \left\{ \nabla_{W, x} g \right\} . \left\{ \diag(\delta(\nabla_x g)) \right\} \bigg\rvert_{(x_i, W)},
\end{equation}
where $\delta(.)$ is the Dirac delta function. Assume that for the $i$-th training sample, there exists some element $j$ such that: 
\begin{itemize}
    \item $\text{(A1):} ~~~ [\nabla_x g]_j\big\rvert_{(x_i, W)} = 0$, 
    \item $\text{(A2):} ~~~ [\nabla_x g]_j \big\rvert_{(x^\prime_i, W)} \neq 0$, and
    \item $\text{(A3)}: ~ [\nabla_{W, x}g]^{(j)}\big\rvert_{(x_i, W)} \neq {\bf 0}$,
\end{itemize}
where $[a]_j$ and $[B]^{(j)}$ denote the $j$-th element of the vector $a$, and the $j$-th column of the matrix $B$, respectively.
 Under (A1), the $j$-th column of $\nabla_W \eta_i$, denoted as $[\nabla_W \eta_i]^{(j)}$, would be an $\infty$ multiple of a non-zero vector ($[\nabla_{W, x} g]^{(j)}$), according to the Eq. \ref{grad_eta}, assumption (A3), and due to the delta function being infinite at zero. In addition, note that the first term in left-hand side of Eq. \ref{Suff_cond} is a weighted summation of column vectors, each with the size of weights:
\begin{equation} \label{weighted_sum}
    \left\{\nabla_W \eta_i\right\} \bigg\rvert_{(x_i, W)}.\left\{\nabla_x g \right\} \bigg\rvert_{(x^\prime_i, W)} = \sum_k [\nabla_W \eta_i]^{(k)} [\nabla_x g]_k
\end{equation}
Therefore, as $[\nabla_W \eta_i]^{(j)}$ is an $\infty$ multiple of a non-zero vector and $[\nabla_x g]_k \neq 0$, according to (A2), the summation in Eq. \ref{weighted_sum} would contain a multiple of $\infty$.
The term $\nabla_W g$ is the weight update that is made during FGSM adversarial training, and according to the Eq. \ref{Suff_cond} is negative of the weighted sum in Eq. \ref{weighted_sum}. Therefore, once getting close to a local minimum in the weight space, the network may experience a huge weight update, which corresponds to the mentioned $\infty$ multiple of $[\nabla_{W, x} g]^{(j)}$ at $x_i$.


A simple way to break (A1) and (A2) is to force the input gradients change smoothly, which is implicitly achieved in GradAlign. Properly clipping the gradient updates of the network's weights might be a way to prevent catastrophic overfitting from happening. Although we do not investigate gradient clipping in our work, we include some initial results in Appendix~\ref{appendix-c5} for future research. The other possibility to mitigate the issue is to zero out tiny input gradients so that the training would not encounter large weight updates close to the local optima. Also motivated by this explanation, we could design the attack based on various random starts and zeroing elements of the attack that are changing across various random starts. This helps to ensure that $\eta_i$ would probably not change too much after the weight update, that is, $\nabla_W \eta_i$ would probably remain small. We will next empirically validate this insight by some experiments on the CIFAR-10 dataset. 

\subsection{Numerical Results}




\begin{figure}[!ht]
\vskip 0.2in
\begin{center}
\centerline{
\includegraphics[width=\columnwidth]{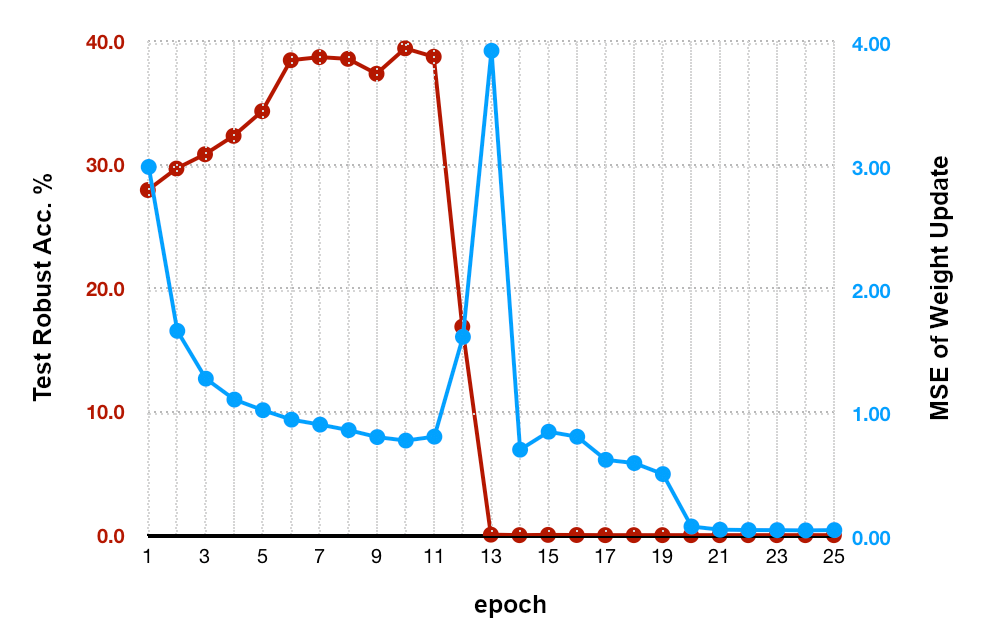}}
\centerline{
\includegraphics[width=\columnwidth]{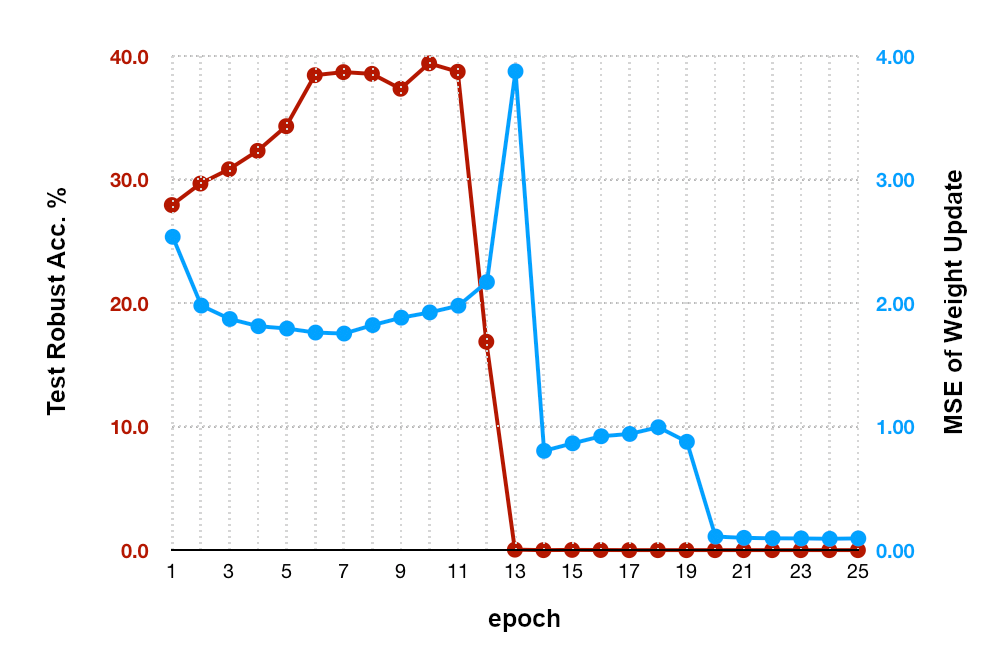}}
\centerline{
\includegraphics[width=\columnwidth]{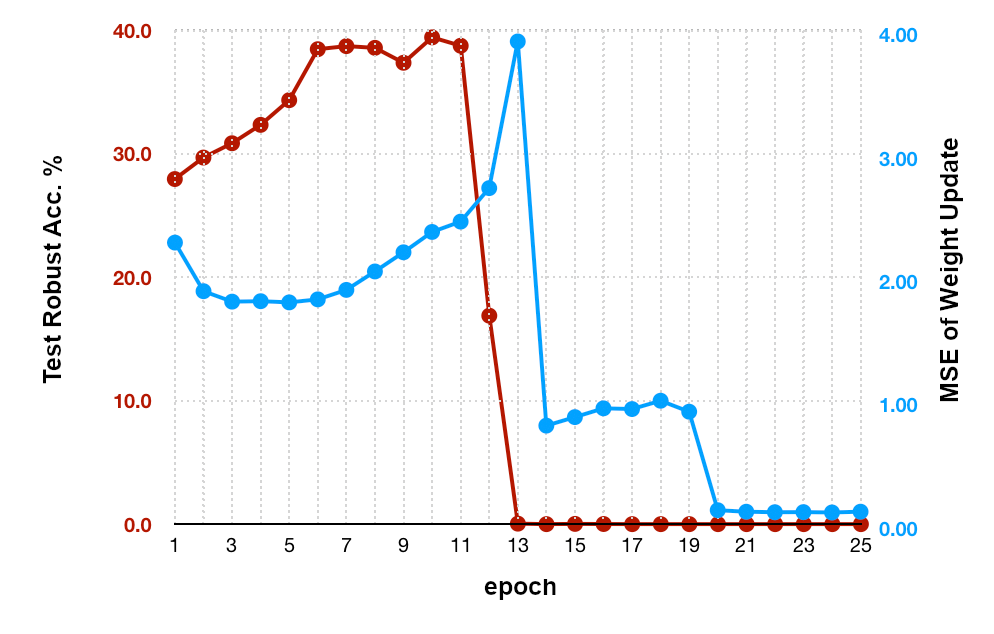}}
\caption{MSE difference of weights, related to three kernels in the convolution layers of the model, before and after updating the model in each epoch (blue), and the test adversarial accuracy of the model (red). The model is trained on the CIFAR-10 dataset with batch size = 128, $\epsilon$ = 8/255, and FGSM-alpha = 2.}
\label{weight_update}
\end{center}
\vskip -0.2in
\end{figure}


Motivated by the explained insights, we tried to observe the trend of a few statistics during the training of a Preact ResNet-18 model on CIFAR-10, specifically exactly at the step when the catastrophic overfitting happens. Here, we use the same default setting for training that is provided in \cite{wong2020fast}. We observe the $\ell_2$ norm of difference in parameters of the model before and after updating the model in an epoch. Interestingly, a huge significant change in parameters is seen exactly at the same epoch that catastrophic overfitting happens. Fig. \ref{weight_update} displays the weight update of three parameters of the network, indicating a huge weight update exactly simultaneous to the drop of test adversarial accuracy of the model.

\begin{figure}[ht]
\vskip 0.2in
\begin{center}

\centerline{
\includegraphics[width=\columnwidth]{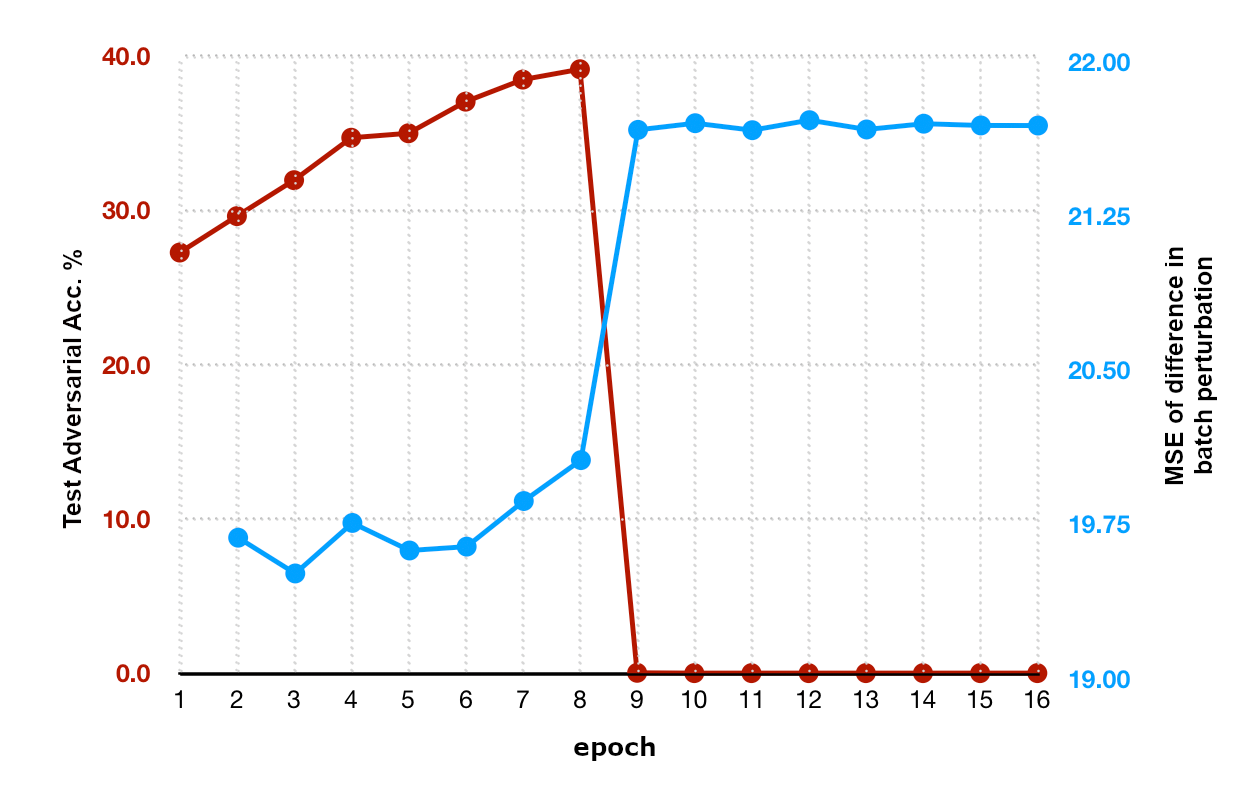}}
\centerline{
\includegraphics[width=\columnwidth]{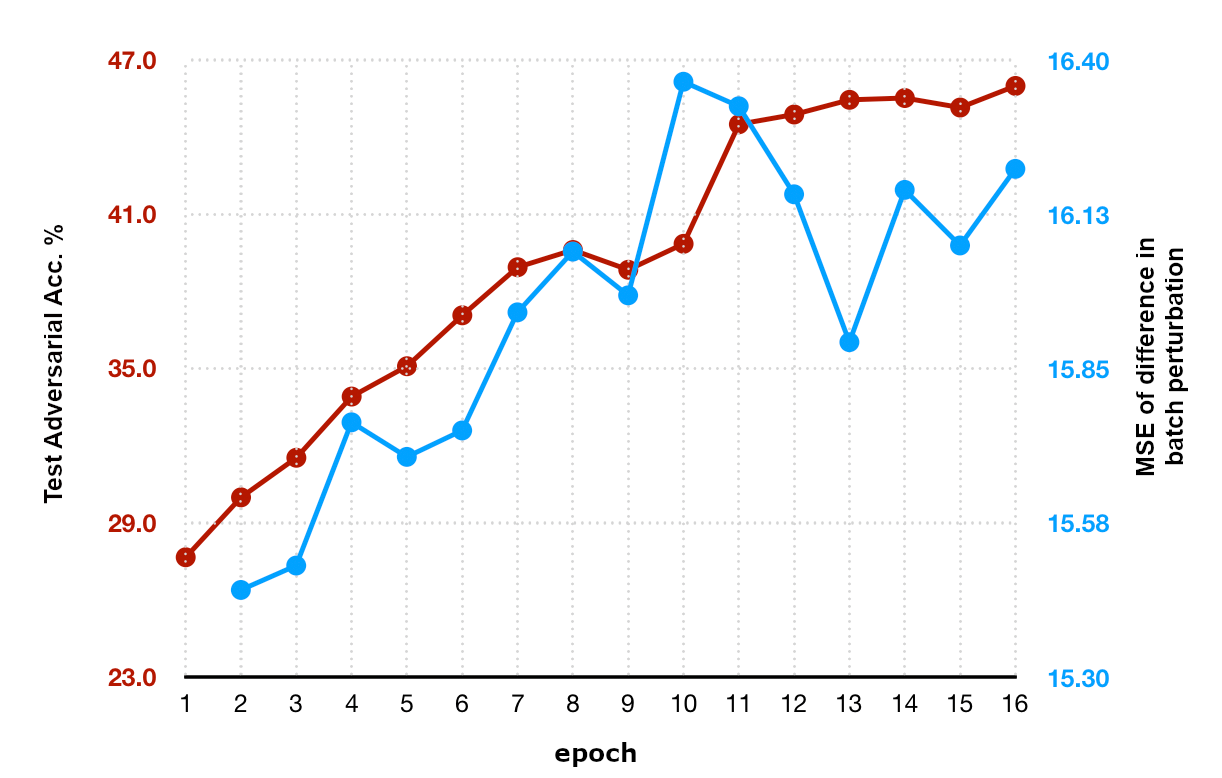}}
\caption{MSE based on the difference in batch perturbations of each two consecutive epochs (blue), and test adversarial accuracy of the model (red). The top diagram refers to training by FGSM-RS. The bottom diagram refers to training by the method that is similar to  FSM-RS, but with zeroing 35\% lower quantile of absolute gradients. The model is trained on the CIFAR-10 dataset with batch size = 128, $\epsilon$ = 8/255, and FGSM-alpha = 2.}


\label{attack_difference}
\end{center}
\vskip -0.2in
\end{figure}

Another observation is the significant change in the FGSM perturbations of a mini-batch before and after catastrophic overfitting, which is shown at top of Fig. \ref{attack_difference}. The jump in the perturbation difference is in accordance with our hypothesis that once the FGSM-trained model gets close to the convergence, the derivative of FGSM perturbations could increase sharply. Due to the idea that tiny elements of the gradient bring about the drastic change of the model, we try to verify that whether ignoring small gradients in FGSM could change this observation or not. To achieve this, we test the difference between FGSM perturbations of a mini-batch in two consecutive epochs when zeroing tiny elements of the input gradient. Specifically, we changed the 35\% lower quantile of the absolute input gradients to zero and acquired the perturbation difference by this modified gradient. MSE based on the difference of perturbations of the same mini-batch in this setting is shown at the bottom of Fig. \ref{attack_difference}. Clearly, in this case, the drastic change of perturbations is prevented. This intuition can help us to propose our new methods based on the single-step attack that is potentially safe from catastrophic overfitting.
\section{Proposed Method}
\label{method}
According to the given insights, we expect that huge weight updates be avoided by zeroing the tiny gradients; or instead zeroing elements of attack that are not the same across different random starts. Next, we describe these methods in more detail.

\subsection{ZeroGrad}
Let $\delta$ be the perturbation that is designed based on FGSM. We propose to take $\bar{\delta} := \delta \odot \mathbb{I}(|\nabla_x g| \geq t)$, where $\mathbb{I}(.)$ is the indicator function, $\odot$ is the element-wise product, and $t$ is a threshold. To make selection of $t$ easier, we propose to adapt $t$ to the sample and use the lower $q$-quantile of the absolute value of the input gradient for each sample. We note that as long as $t$ is small, the loss function would not change too much compared to the original FGSM attack:
\begin{equation}
\begin{split}
    g(x+\bar{\delta}) & \approx g(x) + \bar{\delta}^\top \nabla_x g  \\
    & \geq g(x) + \delta^\top \nabla_x g - \epsilon \sum_{|[\nabla_x g]_i| \leq t} | [\nabla_x g]_i | \\
    & \geq \underbrace{ g(x) + \delta^\top \nabla_x}_{\approx g(x+\delta)} - \epsilon d q t,
\end{split}
\end{equation}
where $d$ is the input dimension, and $t$ is assumed to the be the quantile threshold as described earlier. Therefore, the new perturbation $\bar{\delta}$ is still adversarial for small $t$.  By tuning the threshold and zeroing out the lower gradients, we are able to avoid catastrophic overfitting altogether without sacrificing too much accuracy as shown in section \ref{experiments}. The pseudo code for this method is shown in Alg. \ref{alg:zerograd}.

\begin{algorithm}[tb]
   \caption{ZeroGrad}
   \label{alg:zerograd}
\begin{algorithmic}
   \STATE {\bfseries Input:} number of epochs $T$, maximum perturbation $\epsilon$, quantile value $q$, step size $\alpha$, dataset of size $M$, network $f_{\theta}$
   \STATE {\bfseries Output:} Robust network $f_{\theta}$
   \FOR{$t=1$ {\bfseries to} $T$}
   \FOR{$i=1$ {\bfseries to} $M$} 
   \STATE $\delta \sim \Uniform([-\epsilon, \epsilon]^d)$
   \STATE $\nabla_{\delta} \leftarrow \nabla_{\delta} \ell (f_{\theta}(x_i + \delta), y_{i})$
   \STATE $\nabla_{\delta} [\nabla_{\delta} < \quantile(|\nabla_{\delta}|, q)] \leftarrow 0$
   \STATE $\delta \leftarrow \delta + \alpha \cdot \sgn(\nabla_{\delta})$
   \STATE $\delta \leftarrow \max(\min(\delta, \epsilon), -\epsilon)$
   \STATE // Update network parameters with some optimizer
   \STATE $\theta \leftarrow \theta - \nabla_{\theta} \ell  (f_{\theta}(x_i + \delta), y_{i})$
   \ENDFOR
   \ENDFOR
\end{algorithmic}
\end{algorithm}

\subsection{MultiGrad}
We try to tackle the problem of zeroing out \emph{problematic} gradient elements by identifying the \emph{fragile} ones in a different way. While it does seem that these fragile gradients are generally of small magnitude, we have no reason to believe that all gradients of small magnitude are fragile. Thus it would seem that in the process of zeroing out all gradients below a certain threshold, we are eliminating perturbations that are not necessarily contributing to catastrophic overfitting, and losing some robustness as a result.

As is the sake of the name, these fragile gradients easily change their sign as we move in the $\epsilon$ ball around our sample.
Let  $\delta_1, \delta_2, \ldots, \delta_k$ be $k$ different FGSM-RS perturbations corresponding to $k$ random starting positions.
Let $A$ be the set of indices of the input gradient elements where these perturbations \emph{all} agree on the same sign:
\begin{equation}
\begin{split}
    A=\{i | \ 1\leq\ i\ \leq\ d ,\ [\sgn(\delta_{1})]_i \ = ~ & [\sgn(\delta_{j})]_i \ \\ &\forall : 1 < j \leq k\}
\end{split}
\end{equation}
We define $\bar{\delta}^\prime$ as the perturbation that is equal to $\delta_{1}$ for the elements that are corresponding to the indices of $A$, and equal to zero for the rest. Details of the algorithm is given in the Alg. \ref{alg:multigrad}.

\begin{algorithm}[tb]
   \caption{MultiGrad}
   \label{alg:multigrad}
\begin{algorithmic}
   \STATE {\bfseries Input:} number of epochs $T$, maximum perturbation $\epsilon$, number of samples $N$, step size $\alpha$, dataset of size $M$, network $f_{\theta}$
   \STATE {\bfseries Output:} Robust network $f_{\theta}$
   \FOR{$t=1$ {\bfseries to} $T$}
   \FOR{$i=1$ {\bfseries to} $M$}
   \STATE $\nabla_{\delta_{1..N}} \leftarrow {\bf 0}$
   \FOR{$j=1$ {\bfseries to} $N$ (in parallel)} 
   \STATE $\delta_j \sim \Uniform([-\epsilon, \epsilon]^d)$
   \STATE $\nabla_{\delta_j} \leftarrow \nabla_{x} \ell (f_{\theta}(x_i + \delta_j), y_{i})$
   \ENDFOR
   \STATE $\omega \leftarrow \frac{1}{N} \sum_{j=1}^{N} \sgn(\nabla_{\delta_j})$
   \STATE $\nabla_{\delta} \leftarrow 0$
   \STATE $\nabla_{\delta} [|\omega|==1] \leftarrow \nabla_{\delta_1}$
   \STATE $\delta \leftarrow  \alpha \cdot \sgn(\nabla_{\delta})$ 
   \STATE $\delta \leftarrow \max(\min(\delta, \epsilon), -\epsilon)$
   \STATE // Update network parameters with some optimizer
   \STATE $\theta \leftarrow \theta - \nabla_{\theta} \ell  (f_{\theta}(x_i + \delta), y_{i})$
   \ENDFOR
   \ENDFOR
\end{algorithmic}
\end{algorithm}

\section{Experiments}
\label{experiments}
In this section, we demonstrate the effectiveness of our proposed methods on CIFAR-10, CIFAR-100, and SVHN datasets. We compare ZeroGrad and MultiGrad with Fast (FGSM-RS) and GradAlign, based on the standard test accuracy, and robust test accuracy. For each method, the standard deviation and the average of the test accuracies are calculated by training the models using two random seeds.

{\bfseries Attacks and models.} To report the robust accuracy of our models on the test data, we attack the models with the PGD adversarial attack with 50 steps, 10 restarts, and step size $\alpha=2/255$. We also evaluate our methods based on AutoAttack \cite{croce2020reliable} to make sure that our methods do not suffer from the gradient obfuscation (Appendix~\ref{appendix-a}). The network that is used for training is Preact ResNet-18 \cite{he2016identity}. The results for training with WideResNet-34 are also available in the Appendix~\ref{appendix-c1}, which have better accuracies compared to training with Preact ResNet-18, but the training is much slower.

{\bfseries Learning rate schedules.} There are two types of learning rate schedules that are used for our experiments. The first one is the \emph{cyclical learning rate schedule} \cite{smith2017cyclical}, which helps us get a faster convergence with a fewer number of epochs. We set the cyclical learning rate schedule to reach its maximum learning rate when half of the epochs are passed. The other one is \emph{one-drop learning rate schedule}, which starts with a fixed learning rate and decreases it by a factor of 10 in the last few epochs. For example, if we want to train for 52 epochs and the initial learning rate is 0.1, we drop the learning rate to 0.01 in the 50-th epoch. The reason for stopping the training a few epochs after the drop of the learning rate is to prevent the normal overfitting of the model to the training data \cite{rice2020overfitting}. If the training continues after the learning rate drop, the robust training loss keeps on decreasing but the robust test loss would increase. Note that if the training continues after the learning rate drop, catastrophic overfitting does not happen using ZeroGrad with suitable $q$ or MultiGrad with $N=3$. But it results in worse robust test accuracy due to the general overfitting in adversarial training, which is different from catastrophic overfitting. The one-drop learning rate is not used in the experiments reported in this section. But results with this learning rate schedule are provided in Appendix~\ref{appendix-c3}, as it enables the model to achieve better accuracy if it is trained for more epochs.

{\bfseries Setup for our proposed methods.} For the ZeroGrad method, we use different values of $q$ based on the dataset, size of the network, and $\epsilon$. But we always use the FGSM step size $\alpha=2.0$, which is the maximum possible step size for this method. For the MultiGrad method, we usually use three random samples ($N=3$) because it seems to be enough in most cases. The step size that we use for the MultiGrad method is $\alpha=1.0$. This is the maximum possible step size because in this method we add the final perturbation to the clean sample itself, unlike the ZeroGrad method that starts from a random point in the $\epsilon$ $\ell_\infty$-ball around the clean sample. More analysis on hyperparameters of our methods are available in Appendix~\ref{appendix-b}.

\begin{table}[t]
\caption{Standard and PGD-50 accuracy on the CIFAR-10 dataset with $\epsilon=8/255$ for different training methods. All models are trained with the cyclical learning rate schedule for 30 epochs.}
\label{cifar10-ep8-table}
\vskip 0.15in
\begin{center}
\begin{small}
\begin{sc}
\begin{tabular}{lcccr}
\toprule
Method & Standard Acc. & PGD-50 Acc.\\
\midrule
ZeroGrad (q=0.35)           & 81.61$\pm$ 0.24& 47.55$\pm$ 0.05\\
MultiGrad (N=3)             & 81.38$\pm$ 0.30& {\bfseries 47.85$\pm$ 0.28}\\
FGSM-RS ($\alpha=1.25$)     & 84.32$\pm$ 0.08& 45.10$\pm$ 0.56\\
GradAlign                   & 81.00$\pm$ 0.37& 47.58$\pm$ 0.24\\
\midrule
PGD-2                   & 82.15$\pm$ 0.48& 48.43$\pm$ 0.40\\
PGD-10                  & 81.88$\pm$ 0.37& {\bfseries 50.04$\pm$ 0.79}\\
\bottomrule
\end{tabular}
\end{sc}
\end{small}
\end{center}
\vskip -0.1in
\end{table}

\subsection{CIFAR-10 Results}
For the CIFAR-10 dataset, we use the cyclic learning rate with a maximum learning rate of 0.2 and 30 epochs to be able to compare the results of our methods to the accuracies that are reported for the previously proposed methods. The results for maximum perturbation size $\epsilon=8/255$ are shown in Table~\ref{cifar10-ep8-table}. It seems that larger perturbation sizes like $\epsilon=16/255$, are not valid for the CIFAR-10 dataset, and can completely change labels of the perturbed images \cite{tramer2020fundamental}. However, we also investigated our methods on $\epsilon=16/255$ to see how it performs (see Appendix~\ref{appendix-c2}). Note that our methods can be trained for a higher number of epochs. For example, with appropriate settings, we can train our models for 200 epochs with the piecewise learning rate schedule without encountering catastrophic overfitting. The results for training with more epochs, which lead to better accuracies are available in Appendix~\ref{appendix-c3}.

\subsection{CIFAR-100 Results}
To train our models on the CIFAR-100 dataset, we again use the cyclic learning rate with a maximum learning rate of 0.2 and 30 epochs of training. The models are trained with maximum perturbation size $\epsilon=8/255$. As can be seen in Table~\ref{cifar100-ep8-table}, our methods are able to outperform other models that are trained using single-step adversarial attacks. We keep in mind that although the difference between the robust accuracy of our methods and the GradAlign method might not be much, the main contribution of our methods is that they provide simplicity, speed, and high robust accuracy at the same time.

\begin{table}[t]
\caption{Standard and PGD-50 accuracy on the CIFAR-100 dataset with $\epsilon=8/255$ for different training methods. All models are trained with the cyclical learning rate schedule for 30 epochs.}
\label{cifar100-ep8-table}
\vskip 0.15in
\begin{center}
\begin{small}
\begin{sc}
\begin{tabular}{lcccr}
\toprule
Method & Standard Acc. & PGD-50 Acc.\\
\midrule
ZeroGrad (q=0.45)       & 53.70$\pm$ 0.23& 25.08$\pm$ 0.07\\
MultiGrad (N=3)         & 53.32$\pm$ 0.35& {\bfseries  25.31 $\pm$ 0.03}\\
FGSM-RS ($\alpha$=1.25) & 49.33$\pm$ 0.57& 0.00$\pm$ 0.00\\
FGSM-RS   ($\alpha$=1.0)& 56.94$\pm$ 0.25& 23.78$\pm$ 0.41\\
GradAlign               & 51.92$\pm$ 0.18& 24.52$\pm$ 0.10\\
\midrule
PGD-2                    & 52.45$\pm$ 0.12& 26.72$\pm$ 0.02\\
PGD-10                   & 51.29$\pm$ 0.30& {\bfseries  26.79$\pm$ 0.14}\\

\bottomrule
\end{tabular}
\end{sc}
\end{small}
\end{center}
\vskip -0.1in
\end{table}

\subsection{SVHN Results}
For our experiments on SVHN, we had a slightly different approach. As observed before, the SVHN dataset behaves in peculiar ways, and it seems that it requires a higher quantile for ZeroGrad to be able to prevent \emph{catastrophic overfitting}, i.e. $q =0.7$. We hypothesize that this is due to a large portion of the SVHN images not being relevant to the object of interest i.e. the house number, and therefore having much smaller gradients.  Furthermore, a model trained on this dataset reaches relatively high test accuracies early in the training and then begins to get worse over the remaining epochs.

While this phenomenon still remains unexplained to us and could be of interest for further research, it seems that early stopping does not hinder the \emph{maturity} of the model when trained with our methods. Hence we conduct the experiments in the following way. We train the model for 15 epochs using a cyclic learning rate schedule and we record the robust accuracy on a validation set every epoch. Once training is finished, we test the model with the highest recorded validation accuracy with PGD-50 with 10 restarts on the test set and report the results, which are shown in Table 3. 

\begin{table}[t]
\caption{Standard and PGD-50 accuracy on the SVHN dataset with $\epsilon=8/255$ for different training methods. }
\label{SVHN-ep8-table}
\vskip 0.15in
\begin{center}
\begin{small}
\begin{sc}
\begin{tabular}{lcccr}
\toprule
Method & Standard Acc. & PGD-50 Acc.\\
\midrule
ZeroGrad (q=0.7)       & 88.36 $\pm$ 0.63 & 39.42 $\pm$ 1.92 \\
MultiGrad (N=3)         & 90.25$\pm$ 0.74 & {\bfseries  43.66$\pm$1.06}\\
FGSM-RS ($\alpha$=0.875) & 92.25$\pm$ 0.01& 38.73$\pm$ 0.07\\
GradAlign               & 92.36$\pm$ 0.47& 42.08$\pm$ 0.25\\
\midrule
PGD-2                    & 92.68$\pm$ 0.45& 47.28$\pm$ 0.26\\
PGD-10                   & 91.92$\pm$ 0.40& {\bfseries  52.08$\pm$ 0.49}\\

\bottomrule
\end{tabular}
\end{sc}
\end{small}
\end{center}
\vskip -0.1in
\end{table}

\subsection{Training Time}
Without a doubt, one of the main motivations for the FGSM-based attacks is to be fast. Therefore, it is worth mentioning this aspect of our proposed methods.
The running time for one epoch of training with ZeroGrad is almost equal to the FGSM-RS method \cite{wong2020fast}, since no extra computational overhead is needed in this approach, except for calculating the thresholds. For MultiGrad with the number of random samples $N=3$, the running time is double the running time of FGSM-RS. However, the strength of this method is that calculating the gradients for the random samples can be done in parallel, and by using multiple GPUs, we can reduce the training time. Note that this parallelization is not possible while training with multi-step adversarial attacks like PGD. So approaches like \cite{li2020towards} can not reduce the training time as much as FGSM-RS \cite{wong2020fast}. In addition, GradAlign is more than two times slower than FGSM-RS and also cannot be parallelized \cite{andriushchenko2020understanding}.

The half-precision calculation technique \cite{micikevicius2017mixed}, which leads to nearly two times speedup in FGSM training, can also be used to speed up our methods without having a considerable effect on the standard or robust accuracy of our models. The reported results in our experiments are without using the half-precision technique.

The type of GPUs and other system specifications are inconsistent across papers and does not make much sense to compare the reported ones. In our experiments, by using T4 GPU, the training time of different methods for 30 epochs, using half-precision \cite{micikevicius2017mixed} on CIFAR-10 is: FGSM-RS: 22.5, ZeroGrad: 22.6, MultiGrad: 43.9 (non-parallelized), GradAlign: 87.9, PGD-2: 33.8, PGD-10: 124.2 minutes.

\subsection{Perceptually Aligned Gradients by MultiGrad}

\begin{figure}[ht]
\vskip 0.2in
\includegraphics[scale=0.7]{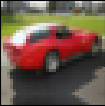} \includegraphics[scale=0.7]{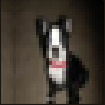}
\includegraphics[scale=0.7]{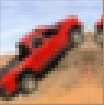} \includegraphics[scale=0.7]{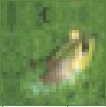}
\\
\\
\includegraphics[scale=0.7]{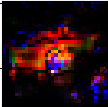} \includegraphics[scale=0.7]{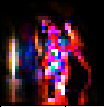}
\includegraphics[scale=0.7]{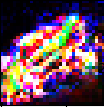} \includegraphics[scale=0.7]{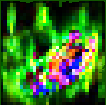}
\\
\\
\includegraphics[scale=0.7]{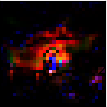} \includegraphics[scale=0.7]{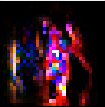}
\includegraphics[scale=0.7]{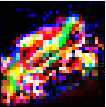} \includegraphics[scale=0.7]{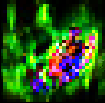}
\\
\caption{
The first row are the input images, the second row are the gradients with respect to the images, and the third row are the gradients that are acquired after applying MultiGrad. As can be seen, the second and third rows are nearly identical and closely related to the foreground object, which shows that Multigrad zeros gradient only in the irrelevant pixels.}
\label{gradient pictures}
\end{figure}

As seen before, adversarially trained models seem to have meaningful gradients with respect to the input image and align well with the human perception \cite{tsipras2018robustness}. The gradient is of larger values on the pixels relating to the object of interest, i.e. the foreground.
We show that after applying the MultiGrad algorithm, the resulting \emph{gradient} preserves this property, which is non-trivial. The results acquired from applying the MultiGrad (N=3) algorithm to various images, pertaining to a Preact ResNet18 model trained on FGSM-RS (before overfitting) can be seen in Fig.~\ref{gradient pictures}. We observed that final gradients acquired from MultiGrad algorithm is also aligned with the perceptually relevant features. This confirms that during this algorithm, valuable gradients have been saved, and it only zeros out fragile gradients, which are mostly non-relevant pixels.

\section{Ablation studies}
In this section, we discuss the effects of hyperparameters that are used in our proposed methods. We mention a rationale and rule of thumb for choosing hyperparameters for a new problem and also introduce a practical technique for adjusting hyperparameters. Finally, we test whether the benefits of the proposed methods originate mainly from lowering the attack $\ell_2$ norm. Additional experiments about the settings of the two suggested methods are available in the Appendix.

\subsection{Hyperparameters selection}
Each of the suggested methods has one main hyperparameter that needs to be adjusted. In ZeroGrad, $q$ is used to determine the threshold for zeroing the gradients based on the lower $q$-quantile of the absolute value of the loss input gradient. The suitable choice of $q$ is related to the given problem and dataset. As mentioned in the previous sections, by zeroing, we are ignoring the small gradients that have less importance, i.e. not relevant to the main object of interest in the sample. This appears to be related to the given task, i.e. what percentage of pixels have no/little information about predicting the output. Therefore, the dataset can give us some hints about the percentage of pixels that are informative. Our estimation about the average percentage of such pixels in the images of our dataset provides us a clue to choose the appropriate $q$. 

This reasoning can also be used in MultiGrad for choosing $N$, the number of random samples. In this regard, a high percentage of irrelevant and less important parts, is a sign for a larger $q$ in ZeroGrad, or the higher number of samples needed to check their agreement on the same sign in MultiGrad. For example, SVHN dataset contains real-world images that have few digits as a house number and a usually large portion as background. Through inspecting the images, we realize that many of the pixels have no considerable information about the class label of the images in this dataset. Confirming this point, we observe in our experiments that for SVHN dataset, larger $q$ and higher $N$ is needed to prevent catastrophic overfitting, especially in comparison with CIFAR-10 and CIFAR-100 datasets.

\subsection{Adjusting the Hyperparameter by Rolling Back}
In addition to the initial estimation of hyperparameters according to the learning problem and the dataset, it is possible to adjust hyperparameters by a rolling back approach. We can choose an approximate $q_0$ in the beginning, and start adversarial training using ZeroGrad with $q_0$. We save checkpoints during training to make rolling back possible. Once faced by the catastrophic overfitting phenomenon, we stop the training, increase the chosen $q$ hyperparameter and adjust it to $q_1$, which is larger than $q_0$. We then continue the training with the ZeroGrad algorithm, resuming from a saved checkpoint at the epoch right before catastrophic overfitting. We tested this idea on CIFAR-10, and successfully trained the model for more than 100 epochs without overfitting, starting from $q_0$ equal to 0.25 and increasing it to 0.45 during the experiment.

\subsection{Mean perturbation size}

One may argue that zeroing certain elements of the input gradients, results in lower $\| \delta \|_2$, and similar results could effectively be obtained by lowering the FGSM step size. We note that lowering the perturbation $\ell_2$ norm has earlier been shown to be helpful in avoiding catastrophic overfitting \cite{andriushchenko2020understanding}.  
To investigate this further, we compare different methods based on their average perturbation $\ell_1$ norm.

We compare our proposed techniques with simply reducing the FGSM step size $\alpha$ to decrease the perturbation size during training on the CIFAR-10 dataset. The results that are reported in Table~\ref{mean-pert-table}, are calculated based on the whole training set samples once at the beginning of the training in the second epoch, and once at the end of the training in 52-nd epoch. All different methods either have a fixed perturbation size during the training, or their perturbation size decreases as the model gets more robust to the adversarial attacks. It can be seen that by reducing the FGSM-RS step size to 1.0, the perturbation size would be less than both of our proposed methods, but its PGD-50 test robust accuracy would be 45.44\%, which is less than what our methods can achieve, which is 47.85\%. Overall, FGSM-RS cannot reach a test robust accuracy close to our methods with any value of the FGSM step size $\alpha$. So what our methods are doing is not simply reducing the perturbation size.

\begin{table}[t]
\caption{$\ell_1$ perturbation norm (multiplied by 255) at the second and the 52-nd epochs during training with one-drop learning rate schedule for 52 epochs. The training is done on the CIFAR-10 dataset with $\epsilon=8/255$, and FGSM step size of 2 for ZeroGrad and 1 for MultiGrad.}
\label{mean-pert-table}
\vskip 0.15in
\begin{center}
\begin{small}
\begin{sc}
\begin{tabular}{lcccr}
\toprule
Method & 2nd epoch & 52nd epoch\\
\midrule
ZeroGrad (q=0.35)       & 6.52 & 6.49\\
MultiGrad (N=3)         & 7.15 & 6.13\\
FGSM-RS ($\alpha=1.0$)  & 5.96 & 5.95\\
FGSM-RS ($\alpha=1.25$) & 6.81 & 6.78\\
GradAlign               & 7.89 & 7.89\\
PGD-2                   & 7.21 & 6.67\\
PGD-10                  & 7.52 & 7.16\\
\bottomrule
\end{tabular}
\end{sc}
\end{small}
\end{center}
\vskip -0.1in
\end{table}

\section{Conclusion and Future Work}
A number of recent work tried to understand FGSM and FGSM-RS adversarial training, its unknown phenomenon, ``catastrophic Overfitting," and also suggest solutions for preventing this from happening. However, existing explanations do not tell any acceptable answer to some questions about some aspects of this phenomenon like its suddenness. Furthermore, the previously suggested solutions usually suffer from the increased training time, contrasting with the aim of less computational cost by using single-step attacks in adversarial training.
In this work, we aimed to present a more comprehensive explanation for catastrophic overfitting. Our hypothesis highlights the role of tiny and fragile input gradients in the training with fast gradient sign method. Based on this intuition, we proposed two simple effective methods. Our empirical results show that our methods prevent catastrophic overfitting, and achieve competitive adversarial accuracy with no significant train-time overhead. A more comprehensive study of why the large weight update is corrupting adversarial robustness remains as a subject of future work. This could potentially result in more useful insights, and lead to further advance single-step attacks for the adversarial training. 

\bibliography{example_paper}

\begin{thebibliography}{16}
\providecommand{\natexlab}[1]{#1}
\providecommand{\url}[1]{\texttt{#1}}
\expandafter\ifx\csname urlstyle\endcsname\relax
  \providecommand{\doi}[1]{doi: #1}\else
  \providecommand{\doi}{doi: \begingroup \urlstyle{rm}\Url}\fi

\bibitem[Andriushchenko \& Flammarion(2020)Andriushchenko and
  Flammarion]{andriushchenko2020understanding}
Andriushchenko, M. and Flammarion, N.
\newblock Understanding and improving fast adversarial training.
\newblock In \emph{NeurIPS}, 2020.

\bibitem[Carlini \& Wagner(2017)Carlini and Wagner]{carlini2017towards}
Carlini, N. and Wagner, D.
\newblock Towards evaluating the robustness of neural networks.
\newblock In \emph{2017 ieee symposium on security and privacy (sp)}, pp.\
  39--57. IEEE, 2017.

\bibitem[Croce \& Hein(2020)Croce and Hein]{croce2020reliable}
Croce, F. and Hein, M.
\newblock Reliable evaluation of adversarial robustness with an ensemble of
  diverse parameter-free attacks.
\newblock In \emph{International Conference on Machine Learning}, pp.\
  2206--2216. PMLR, 2020.

\bibitem[Goodfellow et~al.(2015)Goodfellow, Shlens, and
  Szegedy]{Goodfellow2015ExplainingAH}
Goodfellow, I.~J., Shlens, J., and Szegedy, C.
\newblock Explaining and harnessing adversarial examples.
\newblock \emph{CoRR}, abs/1412.6572, 2015.

\bibitem[He et~al.(2016)He, Zhang, Ren, and Sun]{he2016identity}
He, K., Zhang, X., Ren, S., and Sun, J.
\newblock Identity mappings in deep residual networks.
\newblock In \emph{European conference on computer vision}, pp.\  630--645.
  Springer, 2016.

\bibitem[Kim et~al.(2020)Kim, Lee, and Lee]{kim2020understanding}
Kim, H., Lee, W., and Lee, J.
\newblock Understanding catastrophic overfitting in single-step adversarial
  training.
\newblock \emph{arXiv preprint arXiv:2010.01799}, 2020.

\bibitem[Li et~al.(2020)Li, Wang, Jana, and Carin]{li2020towards}
Li, B., Wang, S., Jana, S., and Carin, L.
\newblock Towards understanding fast adversarial training.
\newblock \emph{arXiv preprint arXiv:2006.03089}, 2020.

\bibitem[Madry et~al.(2018)Madry, Makelov, Schmidt, Tsipras, and
  Vladu]{madry2018}
Madry, A., Makelov, A., Schmidt, L., Tsipras, D., and Vladu, A.
\newblock Towards deep learning models resistant to adversarial attacks.
\newblock In \emph{International Conference on Learning Representation}, 2018.

\bibitem[Micikevicius et~al.(2017)Micikevicius, Narang, Alben, Diamos, Elsen,
  Garcia, Ginsburg, Houston, Kuchaiev, Venkatesh,
  et~al.]{micikevicius2017mixed}
Micikevicius, P., Narang, S., Alben, J., Diamos, G., Elsen, E., Garcia, D.,
  Ginsburg, B., Houston, M., Kuchaiev, O., Venkatesh, G., et~al.
\newblock Mixed precision training.
\newblock \emph{arXiv preprint arXiv:1710.03740}, 2017.

\bibitem[Rice et~al.(2020)Rice, Wong, and Kolter]{rice2020overfitting}
Rice, L., Wong, E., and Kolter, Z.
\newblock Overfitting in adversarially robust deep learning.
\newblock In \emph{International Conference on Machine Learning}, pp.\
  8093--8104. PMLR, 2020.

\bibitem[Smith(2017)]{smith2017cyclical}
Smith, L.~N.
\newblock Cyclical learning rates for training neural networks.
\newblock In \emph{2017 IEEE winter conference on applications of computer
  vision (WACV)}, pp.\  464--472. IEEE, 2017.

\bibitem[Tram{\`e}r et~al.(2020)Tram{\`e}r, Behrmann, Carlini, Papernot, and
  Jacobsen]{tramer2020fundamental}
Tram{\`e}r, F., Behrmann, J., Carlini, N., Papernot, N., and Jacobsen, J.-H.
\newblock Fundamental tradeoffs between invariance and sensitivity to
  adversarial perturbations.
\newblock In \emph{International Conference on Machine Learning}, pp.\
  9561--9571. PMLR, 2020.

\bibitem[Tsipras et~al.(2018)Tsipras, Santurkar, Engstrom, Turner, and
  Madry]{tsipras2018robustness}
Tsipras, D., Santurkar, S., Engstrom, L., Turner, A., and Madry, A.
\newblock Robustness may be at odds with accuracy.
\newblock \emph{arXiv preprint arXiv:1805.12152}, 2018.

\bibitem[Vivek \& Babu(2020{\natexlab{a}})Vivek and
  Babu]{vivek2020regularizers}
Vivek, B. and Babu, R.~V.
\newblock Regularizers for single-step adversarial training.
\newblock \emph{arXiv preprint arXiv:2002.00614}, 2020{\natexlab{a}}.

\bibitem[Vivek \& Babu(2020{\natexlab{b}})Vivek and Babu]{vivek2020single}
Vivek, B. and Babu, R.~V.
\newblock Single-step adversarial training with dropout scheduling.
\newblock In \emph{2020 IEEE/CVF Conference on Computer Vision and Pattern
  Recognition (CVPR)}, pp.\  947--956. IEEE, 2020{\natexlab{b}}.

\bibitem[Wong et~al.(2020)Wong, Rice, and Kolter]{wong2020fast}
Wong, E., Rice, L., and Kolter, J.~Z.
\newblock Fast is better than free: Revisiting adversarial training.
\newblock In \emph{International Conference on Learning Representations}, 2020.

\end{thebibliography}
\bibliographystyle{icml2021}

\clearpage
\newpage

\newpage

\appendix

\section*{Appendix}

\section{Evaluation based on AutoAttack}
\label{appendix-a}

In this section, we report  performance of the models that are trained by our proposed methods against the AutoAttack \cite{croce2020reliable} as a standard robustness evaluation tool.
AutoAttack \cite{croce2020reliable} is a parameter-free adversarial attack, which is an ensemble of a few different attacks. Its goal is to evaluate robustness of models in a reliable manner and identify the defenses that give a wrong impression of robustness. Many earlier proposed defenses resulted in much lower robust accuracy compared to other common attacks that are used for evaluation.

To show that training the network with our methods does not cause gradient masking or gradient obfuscation, we evaluate the models based on the AutoAttack. The results of attacking the models by the AutoAttack on different datasets are shown in Table~\ref{autoattack-t}. On the CIFAR-10 dataset, ZeroGrad with $q=0.35$ and MultiGrad with $N=3$ achieve 43.48\% and 44.39\% accuracy, respectively. These accuracies are reasonable compared to the FGSM-RS with step size $alpha=1.25$, and GradAlign \cite{andriushchenko2020understanding}, with the reported accuracy of 43.21\% and 44.54$\pm$0.24\% respectively.

This shows that our methods are performing well against this attack, and increases our confidence on robustness of the models that are trained with the propoesed methods, ZeroGrad and MultiGrad.

\begin{table}[ht]
\caption{Robust test accuracy against the AutoAttack on the CIFAR-10, CIFAR-100, and SVHN datasets. Models are trained with the cyclic learning rate. For the CIFAR-10 and CIFAR-100 datasets, the number of epochs is set to 30, but for the SVHN dataset, the training is done only for 15 epochs.}
\label{autoattack-t}
\vskip 0.15in
\begin{center}
\begin{small}
\begin{sc}
\begin{tabular}{lcccr}
\toprule
Method & & AutoAttack Acc.\\
\midrule
& CIFAR-10 & \\
\midrule
ZeroGrad (q=0.35)           & & 43.48\\
MultiGrad (N=3)             & & 44.39 \\
\midrule
& CIFAR-100 & \\
\midrule
ZeroGrad (q=0.45)           & & 21.15\\
MultiGrad (N=3)             & & 21.62 \\
\midrule
& SVHN & \\
\midrule
ZeroGrad (q=0.7)           & & 30.07\\
MultiGrad (N=3)             & & 38.20 \\
\bottomrule
\end{tabular}
\end{sc}
\end{small}
\end{center}
\vskip -0.1in
\end{table}

\section{Analysis on hyperparameters}
\label{appendix-b}

In this section, we investigate and discuss effects of the two main hyperparameters of our proposed methods ($q$ in ZeroGrad and $N$ in MultiGrad) in the final results.

\subsection{Different settings for the ZeroGrad and MultiGrad}
\label{appendix-b1}

Results of the ZeroGrad and MultiGrad algorithms with the final selected hyperparameters have previously been reported in the paper. In this section, we report the results of our experiments on CIFAR-10 with different hyperparameters ($q$ and $N$).
Table \ref{cifar10-q} shows the results of ZeroGrad with different $q$, and Table \ref{cifar10-n} shows the same for MultiGrad with different $N$. We used a Preact ResNet-18 \cite{he2016identity}, and 30 epochs of training with the cyclic learning rate schedule similar to our other experiments.

\begin{table}[ht]
\caption{Standard and PGD-50 accuracy on the CIFAR-10 dataset with $\epsilon=8/255$ for different settings of ZeroGrad. All models are trained with the cyclical learning rate schedule for 30 epochs.}
\label{cifar10-q}
\vskip 0.15in
\begin{center}
\begin{small}
\begin{sc}
\begin{tabular}{lcccr}
\toprule
Method & Standard Acc. & PGD-50 Acc.\\
\midrule
ZeroGrad (q=0.3)           & 81.27 & 48.00\\
ZeroGrad (q=0.4)             & 81.70 & 47.31 \\
ZeroGrad (q=0.5)     & 82.28 & 45.98 \\
\bottomrule
\end{tabular}
\end{sc}
\end{small}
\end{center}
\vskip -0.1in
\end{table}

\begin{table}[t]
\caption{Standard and PGD-50 accuracy on the CIFAR-10 dataset with $\epsilon=8/255$ for different settings of MultiGrad. All models are trained with the cyclical learning rate schedule for 30 epochs.}
\label{cifar10-n}
\vskip 0.15in
\begin{center}
\begin{small}
\begin{sc}
\begin{tabular}{lcccr}
\toprule
Method & Standard Acc. & PGD-50 Acc.\\
\midrule
MultiGrad (N=2)           & 81.53 & 48.18\\
MultiGrad (N=4)             & 81.79 & 48.31 \\
MultiGrad (N=5)     & 81.87 & 47.87 \\
\bottomrule
\end{tabular}
\end{sc}
\end{small}
\end{center}
\vskip -0.1in
\end{table}

Finding the most appropriate value of $q$ for training the model with ZeroGrad can be challenging in some cases. We note that it depends on different aspects of the model and dataset.  We noted earlier in the paper that the elements of the gradient that do not contribute to catastrophic overfitting are seemingly the ones that are correlated with the main object of interest, i.e. the foreground. Furthermore, it seems that most of the \emph{fragile} gradient elements are small in magnitude. However, we have no reason to belive that all small elements of the gradient are fragile. By zeroing out gradient elements that are not fragile, we may lose some helpful information, and consequently achieve a lower accuracy than what we could get otherwise. 

The SVHN dataset seems to require a higher value of $q$ (i.e. about 70\%) in order to prevent catastrophic overfitting, and even with the higher $q$, ZeroGrad training results in less than the desired accuracy. After a specific point in the training, its accuracy decreases in each epoch. For this reason, we used the validation-based early stopping scheme. It is worth mentioning that this was not the case with the MultiGrad training. MultiGrad on SVHN seems to reach similar results to GradAlign and also its accuracy did not seem to diminish significantly over time. Regardless, we reported the results using the early stopping scheme for the sake of consistency, and fair comparison. The peculiarity of this dataset and its relation to the value of $q$ is still largely unexplained to us aside from the initial hypothesis that we mentioned earlier, and could be  subject of further research.

For the case of the CIFAR-10 dataset, it seems that the lower the value of $q$, the better the model seems to perform, provided that $q$ is sufficiently large to avoid catastrophic overfitting. Hence it seems that the \emph{fragile} elements of the gradient are smaller than the rest, i.e. they make up the lower portion of all gradient elements. The results of training on different values of $q$ for CIFAR-10 can be seen in Table \ref{cifar10-q}. We trained the models for 30 epochs with a cyclical learning rate similar to our other experiments.

\subsection{Adjusting hyperparameters by Rolling Back}
\label{appendix-b2}

We have suggested the \emph{rolling back} technique for adjusting the hyperparameter $q$ in the ZeroGrad. Using this technique, we can update our initial estimation ($q_0$) for the given learning problem. Using a validation dataset for detection of catastrophic overfitting, we can start the ZeroGrad training algorithm with a small $q_0$, until  catastrophic overfitting occurs. The occurrence of this failure mode shows that larger q is needed to keep on training for more epochs in our problem.

We have tested the \emph{rolling back} for CIFAR-10 and CIFAR-100 (with the maximum perturbation $\ell_\infty$ norm of $8/255$), and successfully trained the model for 100 epochs without any failures. 
For both datasets, we started from $q_0 = 0.25$ and by occurrence of the catastrophic overfitting, we increase $q$ of the algorithm to $0.45$. By this approach, we can get an appropriate $q$ and continue the training for a desired number of epochs. Actually, one of the main consequences of catastrophic overfitting is stopping the training procedure prematurely, i.e. before achieving the best possible performance using that setting. Although employing remedies like early stopping prevents from resulting in models with 0 robust accuracy, it is not able to let premature models continue the training and achieve their best potential results. 
Using our approach, we can prolong training of the until we get the best results. During the rolling back technique for CIFAR-10, the best achieved test robust accuracy is 42.27\%.

Furthermore, we also test this technique for training with a larger maximum perturbation size $16/255$ for CIFAR-10. However, as this setting of training has its specific considerations, which will be discussed more in the Appendix~\ref{appendix-c2}, we continue the training for 100 epochs. The test accuracy of the best saved checkpoint during training for $\epsilon=16/255$ and $\alpha = 2$ against PGD-50 is 19.34\%. Observing the behavior of the model during training can give us information about the robustness potential of the model and the settings of the problem. Note that, preventing catastrophic overfitting does not prevent overfitting due to the overtraining of models. Thus we may observe decreasing the test accuracy with a gentle slope. At the final epoch of training CIFAR-100 using the roll back technique, test robust accuracy of the model against PGD-50 was 17.00\%.

\section{Additional experiments}
\label{appendix-c}

\subsection{Training with WideResNet}
\label{appendix-c1}

WideResNet is a network that is more complex than PreActResNet. The goal of this paper is to investigate the catastrophic overfitting in FGSM training and the advantage of FGSM to other methods is its speed. Hence, using a more complex network causes the training to take more time and is not aligned with our goals. But training with different networks might help with better understanding of the methods, and possibly achieving a higher accuracy. 

Here, we use WideResNet-34 with the width factor 10. Training with this network is around 5 times slower than that of the Preact ResNet-18. The accuracy for training with this network is shown in Table~\ref{wide-t}. As expected, both the standard and robust accuracy of the models are higher than those of the Preact ResNet-18. We note that a larger $q$ is needed to prevent the wider architectures from catastrophic overfitting. We hypothesize that such networks are more vulnerable to get catastrophically overfitted by a large weight update, as they contain more filters. 

\begin{table}[t]
\caption{Standard and PGD-50 accuracy with WideResNet-34 on CIFAR-10 datasets. The training is done with the onedrop learning rate schedule and for 52 epochs.}
\label{wide-t}
\vskip 0.15in
\begin{center}
\begin{small}
\begin{sc}
\begin{tabular}{lcccr}
\toprule
Method & Standard Acc. & PGD-50 Acc.\\
\midrule
ZeroGrad (q=0.45)           & 85.84 & 49.20\\
MultiGrad (N=3)             & 85.44 & 49.25 \\
\bottomrule
\end{tabular}
\end{sc}
\end{small}
\end{center}
\vskip -0.1in
\end{table}

\subsection{Training with larger maximum perturbation size $\epsilon$}
\label{appendix-c2}

In this section, we train our models to be able to defend against adversarial attacks with maximum $\ell_\infty$ perturbation norm of $\epsilon=16/255$ on the CIFAR-10 dataset. We believe that high perturbation size might not be a good measurement to compare different methods, as it enables the adversarial attacks to change the real class of some images. But there is no harm in reporting the performance of our methods using higher perturbation sizes.

The MultiGrad method is not able to prevent the catastrophic overfitting with the maximum step size $\alpha=1.0$. It might be possible to use this method for $\epsilon=16/255$ by using a smaller step size, combining this method with ZeroGrad, and maybe use a random start variation of this attack, but we leave these to the future work. The ZeroGrad method is able to prevent catastrophic overfitting using different settings. But in order to use this method, we have to decrease the step size. For example with step size $\alpha=1.5$ and $q=0.6$, ZeroGrad is able to reach PGD-50 robust accuracy of 19.29\%, which is better than FGSM-RS. For $\epsilon=16/255$, FGSM-RS overfits with the step size $\alpha=0.875$, but is able to reach accuracy of $16.73$ with step size $\alpha=0.75$.

For the large maximum perturbation size, our methods are not able to outperform GradAlign that reaches a reported accuracy of 28.88\%. On the other hand, for this epsilon, training with the standard PGD-2 sometimes results in a poor robust test accuracy. But by combining PGD-2 and ZeroGrad, we can prevent this from happening and reach about 28\% accuracy, similar to the reported one by GradAlign. Note that combining PGD-2 and ZeroGrad means using two steps in ZeroGrad, and this still has a computational time advantage against the GradAlign method.

\subsection{Training with more epochs}
\label{appendix-c3}

In this section, we train our models with more epochs on the CIFAR-10 dataset. In order to do so, we use the onedrop learning rate schedule. The ZeroGrad method is able to train with $q=0.35$, and for 52 epochs, and it reaches the robust accuracy of 47.79$\pm$0.11\% against PGD-50. But if we want to train the model with ZeroGrad for 100 epochs or more, we have to increase $q$ in order to prevent catastrophic overfitting, and we would end up with a worse robust accuracy. For example, if we want to train for 102 epochs with $q=0.45$, the robust accuracy would be 47.21\%, which is less than the training for 52 epochs with $q=0.35$.

The MultiGrad method with $N=3$ is able to train for up to 200 epochs with onedrop learning rate. The standard and robust accuracy for training MultiGrad with 52, 102, 152, and 202 epochs with the onedrop learning rate is available in Table~\ref{multigrad-highep-t}. Training with more epochs results in a slightly better robust accuracy and sometimes a better standard accuracy too. It might be more efficient to continue the training for a few more epochs after the learning rate drop, specially while training for high number of epochs before the learning rate drop. Because this may result in a much better standard accuracy, and sometimes better or the same robust accuracy. But here we only continue the training for two further epochs after the learning rate drop, without considering the total number of training epochs.

\begin{table}[ht]
\caption{Standard and PGD-50 accuracy on the CIFAR-10 dataset for MultiGrad that is trained with high number of epochs with onedrop learning rate schedule.}
\label{multigrad-highep-t}
\vskip 0.15in
\begin{center}
\begin{small}
\begin{sc}
\begin{tabular}{lcccr}
\toprule
Epochs & Standard Acc. & PGD-50 Acc.\\
\midrule
52          & 81.34 & 48.06\\
102         & 83.27 & 48.29 \\
152         & 82.85 & 48.43 \\
202         & 83.09 & 48.98\\
\bottomrule
\end{tabular}
\end{sc}
\end{small}
\end{center}
\vskip -0.1in
\end{table}


\subsection{Similarity to PGD perturbation}
\label{appendix-c4}

To further our understanding, we evaluate the similarity of our methods with the Projected Gradient Descent, and compare it with FGSM. In order to do so, we take a look at the mean percentage of difference in sign of the PGD-10 and other methods perturbations over one epoch. That is, the number of non-zero perturbation features that are of different signs with the PGD perturbation divided by the number of all of the features, and averaged over all mini-batches is calculated.
We do so for the model that is trained based on the FGSM (without random starts for any of the methods) and report the values for one epoch before, and one epoch after the catastrophic overfitting occurs. The results are shown in Table \ref{pgdsim}.

It seems that, overall, our methods are reducing the difference in perturbation with the PGD, which is almost expected for ZeroGrad, because all it does is zeroing out elements. However, this also happens in MultiGrad. Furthermore, after overfitting occurs, the difference increases significantly for both FGSM and ZeroGrad, but not by much for the MultiGrad. This goes to show that MultiGrad is in some ways similar to the PGD-10 as the catastrophic change to the model does not seem to increase their difference as is seen with the other methods.

\begin{table}[ht]
\caption{Percentage of gradient elements that differ in the perturbation sign with PGD-10.}
\label{pgdsim}
\vskip 0.15in
\begin{center}
\begin{small}
\begin{sc}
\begin{tabular}{lcccr}
\toprule
Method &  Before Ovft. & after Ovft.\\
\midrule
FGSM            & 17.58 & 48.53\\
MultiGrad (N=3)           & 11.48 & 12.82 \\
ZeroGrad (q=0.4)     & 6.49 & 28.66 \\
\bottomrule
\end{tabular}
\end{sc}
\end{small}
\end{center}
\vskip -0.1in
\end{table}

\begin{table*}[t]
\centering
\caption{Results for FGSM-RS adversarial training combined with gradient clipping on CIFAR-10 dataset with $\alpha=2\epsilon$ for $\epsilon=8/255$. Training is done using cyclical learning rate for 30 epochs and with 3 different random seeds.}
\label{gradclip}
\vskip 0.15in
\begin{center}
\begin{small}
\begin{sc}
\begin{tabular}{lcccr}
\toprule
Clipping Value &  Standard Acc. & PGD-50 Acc. & Max PGD-50 Acc.\\
\midrule
0.02            & 46.67 $\pm$ 0.28 & 35.11 $\pm$ 0.04 & 35.16\\
0.06            & 69.04 $\pm$ 0.56 & 45.48 $\pm$ 0.17 & 45.67 \\
0.10            & 78.23 $\pm$ 2.08 & 31.11 $\pm$ 22.00 & 48.51 \\
\bottomrule
\end{tabular}
\end{sc}
\end{small}
\end{center}
\vskip -0.1in
\end{table*}

\subsection{Gradient Clipping}
\label{appendix-c5}

As stated in Section~\ref{section-2}, large gradient updates lead the model to overfit on FGSM and perform poorly on stronger attacks. To avoid these updates, we can clip them such that they do not interfere with the gainful updates that help with the model's training. While increasing the clipping threshold reduces its effect and might not prevent catastrophic overfitting, decreasing this threshold interrupts the model's convergence. As shown in Table~\ref{gradclip}, with a clipping value of $0.1$, the model can reach a high accuracy compared to the existing methods, but the training process is unstable and the model would overfit in some cases. However, we believe that it might be possible to use a more complex clipping method to achieve stable training with competitive accuracies to our proposed methods. Note that the time overhead of using gradient clipping is negligible compared to the total FGSM training time.






\nocite{langley00}



\end{document}